\documentclass[journal]{IEEEtran}
\usepackage{amsmath, amsfonts} 
\usepackage{graphicx}
\usepackage{float}
\usepackage{array}
\usepackage{booktabs}
\usepackage{multirow}
\usepackage{enumitem}

\usepackage{xcolor}

\usepackage{hyperref}
\newcolumntype{M}{>{\centering\arraybackslash}m{1.25cm}}
\newcolumntype{A}{>{\centering\arraybackslash}m{1.15cm}}
\newcolumntype{L}{>{\centering\arraybackslash}m{1.50cm}}

\begin{document}
\title{Image-Guided Outdoor LiDAR Perception Quality Assessment for Autonomous Driving}
\author{Ce Zhang,~\IEEEmembership{Member, IEEE}, Azim Eskandarian,~\IEEEmembership{Fellow, IEEE}
\thanks{\textbf{Ce Zhang} is a Senior Machine Learning Engineer at Motional AD. 

\textbf{Azim Eskandarian} is the Alice T. and William H. Goodwin Jr. Dean of Engineering at Virginia Commonwealth University}}

\markboth{Journal of \LaTeX\ Class Files,~Vol.~14, No.~8, April~2024}%
{Shell \MakeLowercase{\textit{et al.}}: A Sample Article Using IEEEtran.cls for IEEE Journals}


\maketitle

\begin{abstract}
LiDAR is one of the most crucial sensors for autonomous vehicle perception. 
However, current LiDAR-based point cloud perception algorithms lack comprehensive and rigorous LiDAR quality assessment methods, leading to uncertainty in detection performance. 
Additionally, existing point cloud quality assessment algorithms are predominantly designed for indoor environments or single-object scenarios.
In this paper, we introduce a novel image-guided point cloud quality assessment algorithm for outdoor autonomous driving environments, named the Image-Guided Outdoor Point Cloud Quality Assessment (IGO-PQA) algorithm. 
Our proposed algorithm comprises two main components.
The first component is the IGO-PQA generation algorithm, which leverages point cloud data, corresponding RGB surrounding view images, and agent objects' ground truth annotations to generate an overall quality score for a single-frame LiDAR-based point cloud. 
The second component is a transformer-based IGO-PQA regression algorithm for no-reference outdoor point cloud quality assessment. 
This regression algorithm allows for the direct prediction of IGO-PQA scores in an online manner, without requiring image data and object ground truth annotations.
We evaluate our proposed algorithm using the nuScenes and Waymo open datasets. 
The IGO-PQA generation algorithm provides consistent and reasonable perception quality indices. Furthermore, our proposed IGO-PQA regression algorithm achieves a Pearson Linear Correlation Coefficient (PLCC) of 0.86 on the nuScenes dataset and 0.97 on the Waymo dataset.
\end{abstract}


\section{Introduction}

\renewcommand{\labelenumii}{\arabic{enumi}.\arabic{enumii}}
\renewcommand{\labelenumiii}{\arabic{enumi}.\arabic{enumii}.\arabic{enumiii}}
\renewcommand{\labelenumiv}{\arabic{enumi}.\arabic{enumii}.\arabic{enumiii}.\arabic{enumiv}}

Autonomous vehicles (AVs) have undergone rapid development in recent decades, with perception emerging as a critical and foundational module \cite{8936542, Mehr2022XCARAE, 10208208}. 
Among the suite of sensors employed, the LiDAR sensor stands out as pivotal for AV perception, offering precise 3D spatial features for tasks such as 3D object detection, segmentation, and object velocity estimation, \cite{Singh2023EndtoendAD, Yan2023CrossMT, Zhao2022NumberAdaptivePL, Yang2023LabelFormerOT}.

Although numerous LiDAR-based perception algorithms and robust evaluation metrics exist \cite{Weng2023JointMM, Henderson2016EndtoEndTO, Luiten2020HOTAAH}, the analysis of the input point cloud quality is often overlooked. 
Such analysis is crucial for real-world autonomous driving, where the reliability and precision of a LiDAR perception system depend not only on the perception algorithms but also on the quality of the LiDAR point clouds. 
Factors such as object occlusions and sparsity significantly impact LiDAR point cloud quality \cite{luo2024towards}.
Both industry and academia have proposed various solutions to enhance LiDAR point cloud quality, including stacking multiple high-beam LiDAR sensors, point cloud upsampling, and data augmentations \cite{wang2021pointaugmenting, 10005270, He2023GradPUAP, yuan2024ad, 10159515}.
However, these solutions lack comprehensive quantitative analyses of point cloud quality assessment in autonomous driving environments.

To address this research gap, we propose a methodology to quantitatively analyze point cloud quality for autonomous driving. 
Given the sparse nature of point clouds, we leverage image data to assist in point cloud quality assessment. 

Before detailing our approach, it is essential to distinguish between existing studies and our work.
Firstly, concerning single-object point cloud quality assessment, although there are existing studies in this area, we consider current algorithms unsuitable for evaluating outdoor LiDAR point clouds. 
Most existing algorithms focus on single objects rather than full scenes, which include backgrounds and multiple objects, and they often rely on a reference point cloud as ground truth—something typically unavailable for outdoor real-world LiDAR point clouds \cite{9878979, zhang2022mm, 10403987, 10422856}. 
Secondly, regarding confidence score thresholds from most object detection algorithms, while many 3D object detection methods predict object confidence scores to represent the probability of an object's existence at a specific position, these confidence scores are often misconstrued as indicators of point cloud quality. 
In reality, these scores only reflect the confidence or probability of the specific object's existence.
Aggregating the confidence scores of all detected objects in a scene to assess point cloud quality is not feasible due to the random variation in the number of objects in each scene.

After a thorough investigation of these issues and careful analysis of potential solutions, we propose an image-guided outdoor LiDAR point cloud quality assessment (IGO-PQA) algorithm. 
This algorithm leverages an image-based saliency mapping technique to capture saliency intensities within a given point cloud scene's corresponding images. 
Subsequently, we align the LiDAR-based point cloud with the corresponding images and extract saliency intensities. 
These saliency intensity scores are then integrated with the point cloud's distance to the ego vehicle and ground truth object annotations to derive the point cloud quality score. 
Given that obtaining ground truth annotations and generating saliency intensities can be time-consuming, we develop a transformer-based neural network model capable of directly online predicting the point cloud quality score in real-time. 

The contributions of this paper can be summarized as follows:

\begin{enumerate}
\item We design and validate a novel image-guided LiDAR point cloud quality assessment algorithm. According to the best of our knowledge, this is by far the first study that focused on outdoor LiDAR-based point cloud quality assessment.
\item We develop a transformer-based neural network model to directly predict the point cloud quality score in an online manner.
\item We conduct thorough ablation studies to verify our proposed neural network models' robustness and correctness.
\item We validate our point cloud quality assessment algorithm using two large-scale autonomous driving datasets, namely the nuScenes dataset and the Waymo open dataset.
\end{enumerate}
\section{Related Works}
In this section, we provide a concise overview of various 3D object detection algorithms that utilize LiDAR point cloud data, as we leverage them to assist in evaluating the point cloud quality assessment score. Furthermore, to enhance our understanding of point cloud quality assessment, we present recent studies relevant to this topic.

\subsection{3D Object Detection Algorithm}
3D object detection \cite{li2022bevformer, 10203376, Lin2023Sparse4DVA, 2305.14018} is one of the key components for AVs' perception, as it provides crucial surrounding agent objects' information. 
A typical 3D object detector usually predicts the agent objects' position (${x}$, ${y}$, ${z}$), shape (width (${w}$), length (${l}$), height (${h}$)), orientation (yaw (${\varphi}$)), velocity (${v_x}$, ${v_y}$), and type of classes. 
To achieve optimal detection performance, a lot of 3D object detectors utilize LiDAR point cloud as the input, as it provides precise spatial information. 
Current LiDAR-based 3D object detector leverages neural network algorithms, which can be categorized as CNN-based and transformer-based algorithms.
\subsubsection{CNN-based 3D Detector}
CNNs were first applied in image classification and detection for feature extraction, multi-layer feature aggregation, and anchor box proposals due to their naturally designed 2D spatial convolutions. 
Most CNN-based 3D detectors follow a similar pipeline as 2D detectors. 
The key innovations for most CNN 3D detectors are focused on improving feature extraction efficiency and enhancing anchor box initialization to improve bounding box prediction accuracy.

SECOND \cite{s18103337} is one of the most earliest LiDAR point cloud-based 3D object detectors that focuses on improving CNN feature extraction efficiency.
In their study, they borrow a similar model from 2D object detectors which contains a backbone for feature extraction and a region proposal network for potential anchor boxes proposal but they propose a novel sparse 3D CNN that dramatically improves the 3D detectors' inference and training speed.
Besides speed improvement, they also focused on the 3D bounding box orientation performance improvement by introducing a new loss function. 
Besides SECOND, PV-RCNN \cite{Shi2021PVRCNNPF, Shi2019PVRCNNPF} argue that keypoint features are helpful for 3D detection performances.
They propose a voxel-based set abstraction module to summarize a small set of key points.
Finally, the summarized key points are aggregated and fed into the detection head for 3D detection.
Even though 3D-based voxels can provide better detection accuracy, the computation cost is too high to be deployed on vehicles for real-time inference \cite{Zhou2017VoxelNetEL}. 
Furthermore, since AVs are mainly operated on roads, which can be simplified as 2D surfaces, the idea of pillars is introduced for 3D detection, where replace the z-axis (height) by using a pillar (no explicit z-axis) instead of 3D voxels (explicit z-axis). PointPillars \cite{Lang2018PointPillarsFE}, PillarNet \cite{Shi2022PillarNetRA} and PillarNext \cite{Li2023PillarNeXtRN} are typical examples of such ideas.
In the PointPillar \cite{Lang2018PointPillarsFE} paper, they propose a novel end-to-end 2D convolutional layers to learn features on pillars without utilizing a fixed encoder.
Such an approach significantly improve the 3D detection performance and increase the inference speed as well.
For the PillarNet \cite{Shi2022PillarNetRA}, the authors tried to close the gap between a pillar-based approach and a voxel-based approach. 
In this study, they propose a more powerful point encoder with multi-scale feature extraction and an orientation-decoupled intersection over union (IoU) regression loss. 
The objective of PillarNext \cite{Li2023PillarNeXtRN} is to further close the gap between the voxel-based approach and the pillar-based approach. 
In their study, they argue that the pillar-based approach can achieve even better performance, compared with voxel by enlarging the receptive field.

Besides inference speed and efficiency, recent studies are also focused on 3D box orientation performance. CenterPoint \cite{Yin2020Centerbased3O} proposes a novel center-based object initialization approach to replace the anchor boxes approach, which dramatically improves the box orientation performance.
\subsubsection{Transformer-based 3D Detector}
With the development of neural network models, transformers have become popular due to their scalability and global feature extraction capabilities. 
One of the main applications of transformers in 3D object detection follows the Detection Transformer's (DETR) 2D object detection pipeline \cite{carion2020end, zhu2020deformable}. 
In this approach, queries are initialized as potential object anchors, followed by cross-attention between the queries and the extracted LiDAR feature map. 
Finally, the model predicts the box shape, heading angle, and positions for each query, resulting in the detection bounding box \cite{Bai2022TransFusionRL, Liu2022BEVFusionMM, Chen2023FocalFormer3DF}.
TransFusion \cite{Bai2022TransFusionRL} is one of the early-stage transformer-based detection models for point cloud data. 
This paper adopts a center-based object initialization approach to generate potential object queries. 
It then uses a single-layer transformer decoder to perform self-attention within the proposed object queries and cross-attention between the proposed queries and LiDAR point cloud features, further refining the potential query features for object detection. 
This approach achieved state-of-the-art performance at the time, demonstrating the effectiveness of the DETR-based method for 3D object detection.
The DETR-like model design is robust and flexible to adapt to temporal features.
Therefore, there are studies focusing on improving the detection performance by benefiting the temporal features and even doing simultaneous detection and tracking \cite{9897855, Zhang2023MotionTrackET}.
Besides the DETR-like approach, another transformer application is the point cloud feature aggregation \cite{Chen2023FocalFormer3DF, Huang2023PTTPT, Deng2022VISTAB3}. 
VISTA \cite{Deng2022VISTAB3} is one of the examples that utilize the transformer architecture to combine the birds-eye-view (BEV) and range-view (RV) feature maps.
Besides spatial feature aggregation, temporal aggregation is also popular for 3D object detection.
PTT proposed a trajectory transformer to aggregate temporal point cloud features and box features to refine 3D object detectors through long temporal information.

\subsection{Point Cloud Quality Assessment}
PQA comprises objective PQA and subjective PQA. 
The former evaluates the point cloud data based on examining the point cloud's distortion level, while the latter is summarized by a MOS based on human observations.
Existing PQA algorithms can be categorized as full-reference (FR), reduce-reference (RR), and no-reference (NR) PQA.
FR PQA requires the point cloud data to have original non-distorted point cloud data as a reference, then compare the distorted point cloud with the original point cloud for quality assessment.
RR PQA, similar to FR PQA, also requires original non-distorted point cloud data but only needs a certain portion.
NR PQA, which is more practical for real-world applications, does not contain a reference point cloud for evaluation, where the quality score can be MOSs based on human evaluation or other features obtained from the point cloud.
Based on the best of the author's knowledge, there is no PQA algorithm or dataset for outdoor LiDAR-based point cloud applications, which are concentrated on indoor environments with a single object.
Therefore, this section briefly discusses the existing indoor-based PQA algorithms for readers' reference.

\subsubsection{Full Reference PQA}
FR PQA utilizes distance functions to measure the degradation between the distorted and original point clouds.
A common distance function is the Euclidean distance \cite{Javaheri2021APJ, 10.1117/12.2322741}. 
D. Tian et al. \cite{8296925} introduce a point-to-surface distance based on the Euclidean distance computation.
According to the experiment results, the proposed metrics are independent of the point cloud size and can better monitor the perceived point cloud quality.
Even though Euclidean distance successfully captures the geometric similarity, it fails to obtain point cloud visual quality after point removal.
Based on such drawbacks, angular similarity computation, Hausdorff, and color statistics computation methodologies are invented.

\subsubsection{Reduced Reference PQA}
The RR PQA utilizes statistic features from the original point cloud to predict the visual quality. 
There are limited RR PQA algorithms.
I. Viola et al. extract geometry, color, and normal vector domain features from a small set of the original point cloud to assess the visual degradation \cite{pub:29557}.
They compare the predicted distortion results with the FR PQA distortion results, proving that RR PQA evaluation metric is effective.
W. Zhou proposes a content-oriented saliency projection method, namely RR-CAP, for RR PQA evaluation \cite{Zhou2023ReducedReferenceQA}.
The experimental results demonstrate that their proposed RR PQA outperforms other quality metrics.

\subsubsection{No Reference PQA}
NN-PQA analyzes the distorted 3D point cloud to obtain a quality score without reference data.
Q. Liu develops the first NN PQA \cite{Su2022NoreferencePC}, namely PQA-Net, to regress the MOS with the given distorted point cloud. 
The experiment results show that PQA-Net can successfully predict the MOS and outperforms other methods.
Z. Zhang utilizes support vector regression (SVR) to predict the NR PQA score \cite{Zhang2021NoReferenceQA}.
They project 3D point clouds to specific color and geometry domains, then employ natural scene statistics (NSS) and signal entropy to extract quality-aware features for classification.
The proposed methods outperform existing NR-PQA and achieve similar performance with FR PQA.

Compared with other computer vision (CV) tasks, such as MOD, 3D reconstruction, and pose estimation, there is limited research literature for point cloud quality evaluation.
Numerous obstacles still exist, including the availability of vast open-source datasets, evaluating multiple-point cloud objects, and conducting outdoor assessments.
\section{Methodology}
\subsection{IGO-PQA Motivation}
The objective of the IGO-PQA algorithm is to enable blind quality assessment for outdoor LiDAR-based point clouds, especially within autonomous driving contexts. 
Unlike many blind image quality assessment methods that rely on human graders and manual annotations, assessing the quality of LiDAR-based point clouds requires alignment with LiDAR-based perception algorithms, such as 3D object detection. 
This necessity arises from the fact that LiDAR-based point clouds are designed for perception algorithms rather than direct human observation.

In LiDAR-based perception algorithms, spatial and semantic information are crucial for achieving optimal detection performance.
Therefore, creating a representation that incorporates both spatial and semantic features is essential for assessing the quality of point cloud perception.
Spatial feature representation is readily obtained from LiDAR-based point cloud data, as each LiDAR point is characterized by its x, y, and z distances to the sensor. 
However, acquiring semantic information is challenging because LiDAR point clouds do not explicitly provide rich semantic representations like cameras. 

T. Zheng et al. introduced a point-based saliency mapping method that computes loss gradients by shifting points towards the point cloud centroid to assess pointwise performance \cite{Zheng2018PointCloudSM}. 
However, this technique faces challenges in outdoor environments with multiple objects. 
One reason is that the loss may not solely reflect a single object's classification performance. 
Another limitation is that the point-shifting algorithm performs effectively on densely packed single objects (typically around 1000 points per object) but struggles with sparse point cloud data (approximately 30 to 40 points per object).
Even though such an approach is not applicable to outdoor point cloud quality assessment, we value the saliency mapping idea is valuable, which motivates us to construct a similar saliency-based mapping to represent the importance of the semantic features.
This inspires us to utilize the concept of saliency intensity to compute saliency values for each point in the outdoor LiDAR-based point cloud.
In addition to T. Zheng's point saliency approach, our prior work on image-based saliency mapping \cite{Zhang2022AQI} inspires us to utilize image saliency intensities as guidance for generating point saliency. 
Compared to LiDAR-based point clouds, images offer extensive semantic representations and are easier to construct, providing more explicit semantic information.

Building upon these prior works, we propose the IGO-PQA algorithm, which leverages image saliency intensities \cite{4270292, FLORES201962, DingZE24} as guidance to enhance semantic representations and utilize LiDAR-based point cloud's spatial information to generate a comprehensive LiDAR-based point cloud perception quality score for autonomous driving scenarios.

\subsection{Image-guided Point Cloud Assessment Algorithm}
\begin{figure*}[ht]
\centering
\includegraphics[width=18cm]{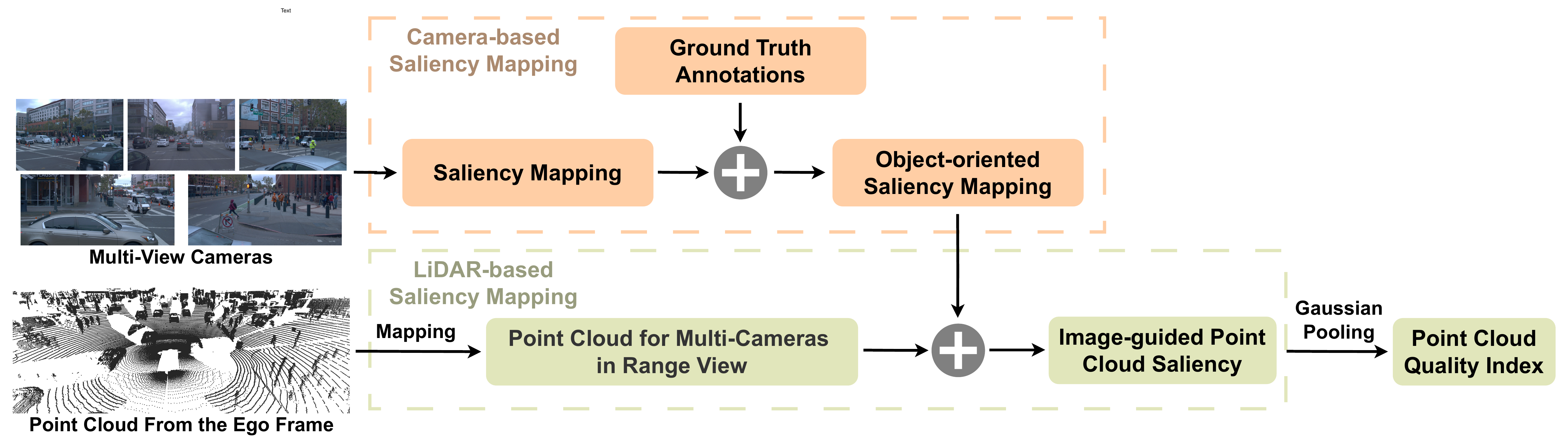}
\caption{Image-guided Point Cloud Quality Index Generation Paradigm}
\label{fig:method_section_overall_pipeline}
\end{figure*}

Figure \ref{fig:method_section_overall_pipeline} illustrates the architecture of the IGO-PQA generation process. 
The IGO-PQA comprises a camera-based saliency mapping module, a LiDAR-based saliency mapping module, and a Gaussian pooling process. 
Initially, RGB images captured by the camera undergo saliency map processing to generate an object-enhanced saliency map. 
Subsequently, the image saliency map guides the projection onto the LiDAR-based point cloud distance saliency map, providing semantic information. 
Then, the camera-guided point cloud saliency map is fed into a Gaussian distribution pooling process to enhance the image-guided saliency intensity. 
Finally, the summation of the overall saliency intensity is normalized across all collected data and considered as the point cloud quality index results.
It is noteworthy that the normalization process simply rescales the summation of saliency intensity to a range of 0 to 100. 
In this study, we emphasize normalizing and rescaling the saliency intensity within different datasets rather than normalizing across all datasets, due to variations in sensor setups. 

\subsubsection{Camera-based Saliency Module}
\begin{figure}[ht]
\centering
\includegraphics[width=8.5cm]{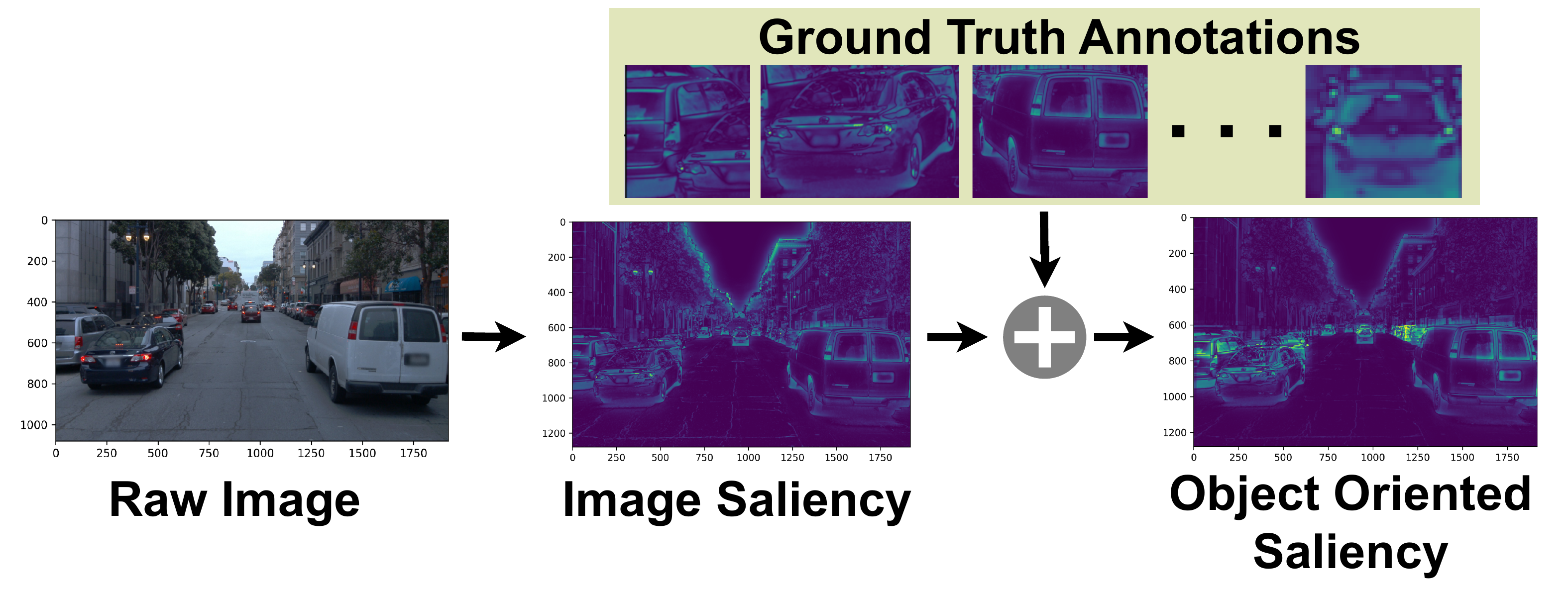}
\caption{Object-orientated Image Saliency Mapping Module for Each Camera}
\label{fig:method_section_image_saliency}
\end{figure}

The camera-based saliency module serves the purpose of extracting meaningful information from camera input. 
The fine-grained saliency map approach distinguishes between object and background characteristics, while the object-oriented saliency map enhances object-specific saliency features. 
Our prior work \cite{Zhang2022AQI} thoroughly explains the fine-grained saliency approach equation, with Figure \ref{fig:method_section_image_saliency} illustrating the process for object-oriented saliency enhancement. The objective of enhancing object-specific saliency is to elevate the saliency intensity value for target objects, emphasizing them in the scene and potentially increasing corresponding point saliency if the point cloud data captures the object. 
The image saliency map resulting from this approach, as shown in the figure, includes background semantic information, such as buildings and vegetation, as well as enhanced object features.  
These outcomes can guide the generation of a saliency map using LiDAR-based point cloud data as a reference.

\subsubsection{LiDAR-based Saliency Module}
\begin{figure}[ht]
\centering
\includegraphics[width=8.5cm]{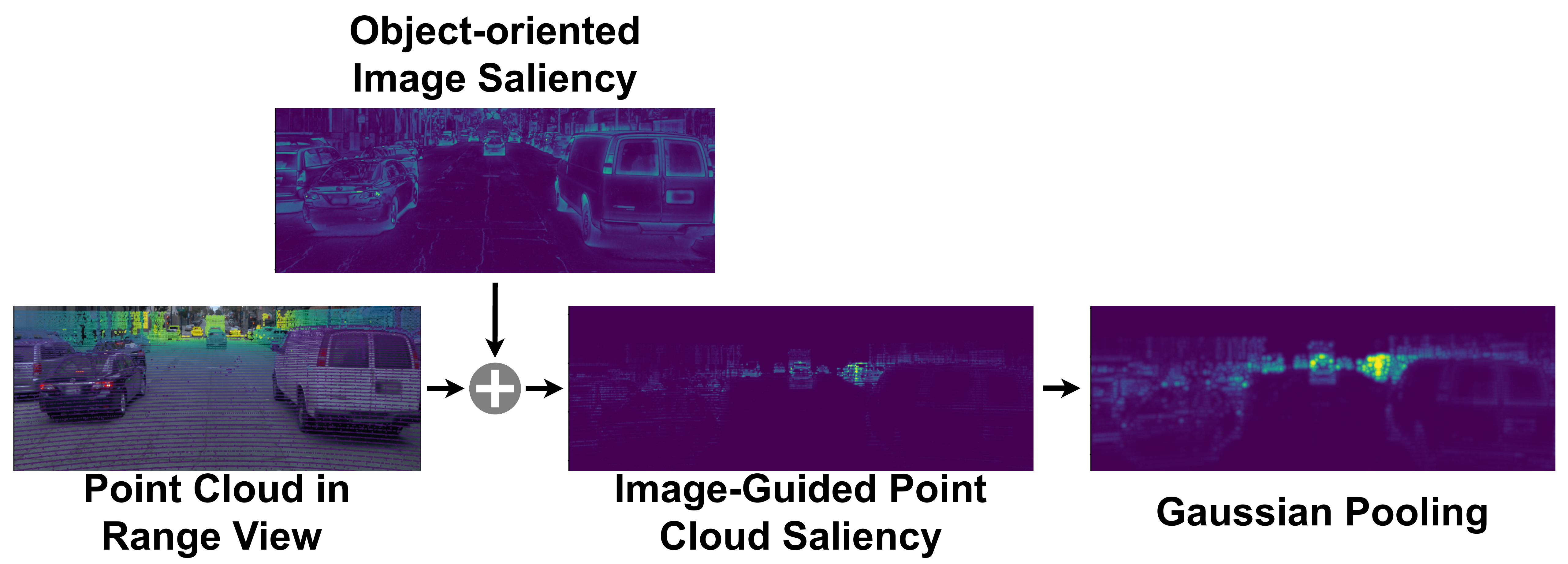}
\caption{Point Cloud Saliency Mapping Module with Image Guidance}
\label{fig:method_section_lidar_saliency}
\end{figure}

The LiDAR-based saliency intensities are derived by integrating semantic and spatial representations.
The spatial representations are obtained directly from the point cloud Cartesian distance, while the semantic representations are guided by image-based saliency intensities, as illustrated in Figure \ref{fig:method_section_lidar_saliency}.
It is worth noting that Figure \ref{fig:method_section_lidar_saliency} demonstrates the LiDAR-based saliency generation module for a single camera.
In our proposed algorithm, we project LiDAR-based point cloud onto all surrounding-view cameras (six cameras for nuScenes and five for the Waymo open dataset) and aggregate the generated point cloud saliency from all cameras.
For ease of explanation, we illustrate the point cloud saliency score computation for one camera.

The process begins by projecting the point cloud data onto the corresponding RGB-based image.
After projection, we utilize the point cloud's Cartesian distance as the spatial representation.
Since distant objects are challenging for LiDAR sensors to capture, objects situated farther away (with higher distances) can receive higher scores, accounting for the reduced point cloud density and smaller appearance of distant objects.
Subsequently, the spatial scores of all point cloud data are normalized to a range of 0 to 1 to align with the intensity range of the image-based saliency map. 
Following this, the image-based saliency map is superimposed onto the distance saliency map. 
Pixel positions of the image saliency intensity that overlap with the distance saliency points are selected to create the image-guided point cloud semantic information saliency map. 
Finally, the normalized distance saliency intensity is multiplied by the image-guided point cloud saliency map to yield the final point cloud saliency intensity map. 
In summary, the overall LiDAR-based saliency mapping module is
\begin{equation}
S_i=\frac{\sqrt{(x_i^2+y_i^2+z_i^2)}}{d}*I_i
\end{equation}
\begin{equation}
S = {S_1, S_2, S_3, ... , S_n}
\end{equation}
where \(S_i\) is the image-guided saliency score for point \(i\), \(x_i, y_i, z_i\) is the Cartesian distance for point \(i\), \(d\) is the maximum distance from all collected point cloud samples, and \(I_i\) is the image saliency intensity that overlap with the point \(i\), and \(S\) is a collection of all image-guided saliency scores for the point cloud projected on an image.
This map integrates both distance (from the point cloud) and semantic information (guided by the RGB-based image), offering a more comprehensive understanding of the environment.

To obtain the 360-degree point cloud saliency score, we simply aggregate (concatenate) all the point cloud saliency scores from the surrounding view images.

\subsubsection{Saliency Pooling Process}
Given that point cloud data is sparser compared to RGB-based image data, each point in the point cloud carries more information than a single pixel in the image, a crucial aspect for LiDAR-based 3D detection algorithms. 
To emulate this characteristic, the saliency pooling process integrates the saliency of neighboring points within a defined radius. 
Here, the process converts a single saliency point into a circle with Gaussian distribution intensity \cite{9709203}. 
This circle delineates the area of influence surrounding the saliency point, with the Gaussian distribution assigning a weight to each neighboring point within the circle based on its distance from the circle's center. Consequently, the outcome is a pooled saliency value that incorporates the impact of multiple neighboring points, as depicted in Figure \ref{fig:method_section_lidar_saliency}.

Ultimately, the pooled saliency map is summed across the surrounding-view cameras to derive the point cloud quality index result (IGO-PQA) for a given frame. 
This index offers a comprehensive evaluation of the overall quality of the point cloud data in the frame, considering both the saliency of individual points and their influence on neighboring points.
As above-mentioned, after generating the IGO-PQA score for a whole dataset, we normalize the generated IGO-PQA score within the dataset to 0~100.

\subsection{IGO-PQA Regression Model}
\begin{figure}[ht]
\centering
\includegraphics[width=8.5cm]{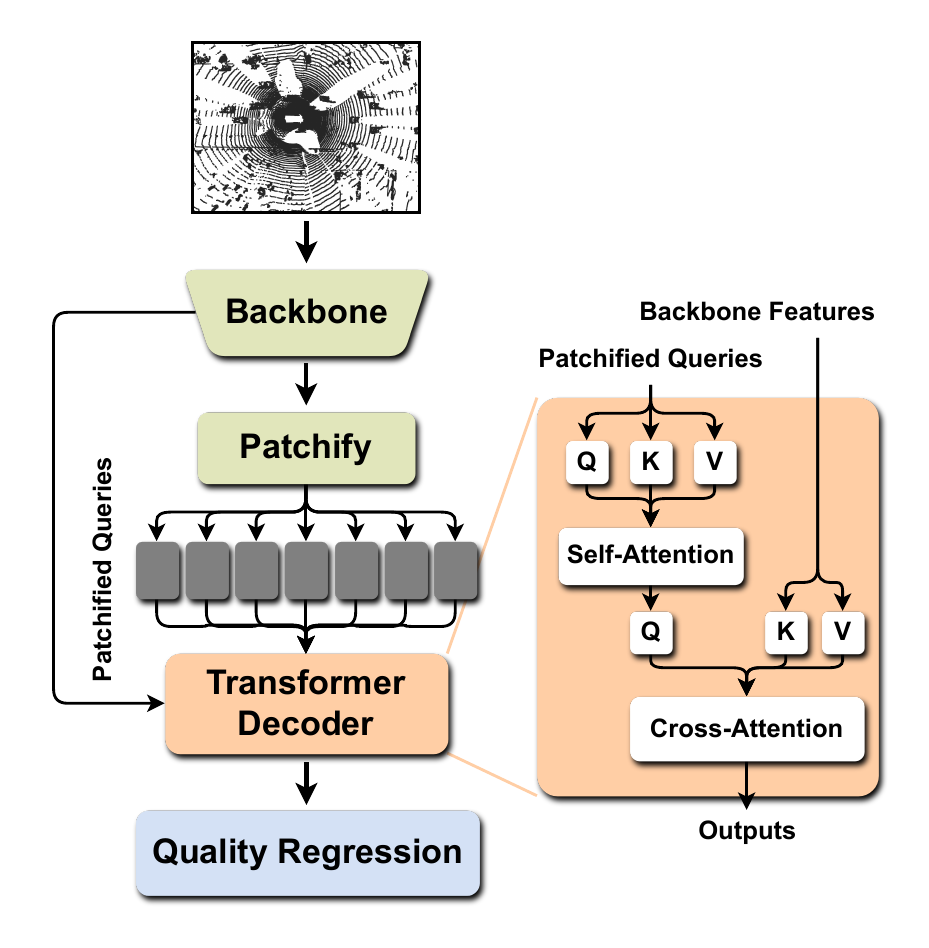}
\caption{Point Cloud Quality Assessment Regression Model}
\label{fig:method_section_regression_model}
\end{figure}

The previously illustrated IGO-PQA generation algorithm provides an index score representing the LiDAR-based point cloud quality in outdoor environments. 
However, the proposed algorithm has several limitations: it requires ground truth annotations, and normalization across datasets, and the saliency mapping process is slow, making it unsuitable for online processing.
To estimate the IGO-PQA score online, a regression model is necessary, which is based on a common outdoor LiDAR-based point cloud backbone for feature extraction and a transformer architecture to aggregate the extracted features for IGO-PQA regression, as illustrated in Figure \ref{fig:method_section_regression_model} 
To the best of our knowledge, there are few existing algorithms for outdoor LiDAR-based point cloud quality assessment in autonomous driving applications, making our proposed IGO-PQA regression model a pioneering benchmark.

The intuition behind the proposed IGO-PQA regression model is that point cloud data quality is determined by factors such as point cloud sparsity (density) and detection range. To capture these aspects, we group the point cloud data into small rectangular patches and extract spatial features within each patch to obtain point cloud sparsity features through a self-attention module. These extracted features are then used in a cross-attention mechanism to analyze the entire LiDAR-based point cloud feature map, allowing the model to learn the range features effectively.

For the IGO-PQA regression model, we apply both PointPillars and VoxelNet as point encoders and backbones for different setups. 
The PointPillars backbone provides faster processing speed but less accurate point cloud features, while the VoxelNet backbone offers higher accuracy at the cost of processing speed. 
Both backbones have proven effective for outdoor autonomous driving applications.

The details of the transformer design are shown in Figure \ref{fig:method_section_regression_model}, which uses query patches and the global query feature map as input, similar to DETR-based transformer architectures. 
Since the IGO-PQA evaluates the overall quality of a given point cloud data frame, we divide the query feature map into patches of the same size and embed them as queries. 
The patched query features are then passed into a transformer encoder for feature extraction to learn the density features within each patch. 
The encoded query features are projected to the point cloud feature map through cross-attention modules, serving as the transformer decoder. 
The aim of this process is to (a) determine which query patches are more influential and (b) learn the point cloud's range features across the whole LiDAR map. 
After the decoding process, the extracted features from each patched query are fed into a series of multi-layer perceptron (MLP) layers for score regression.

\section{Results and Sections}
\subsection{Experiment Dataset}
\begin{figure}[ht]
\centering
\includegraphics[width=8.5cm]{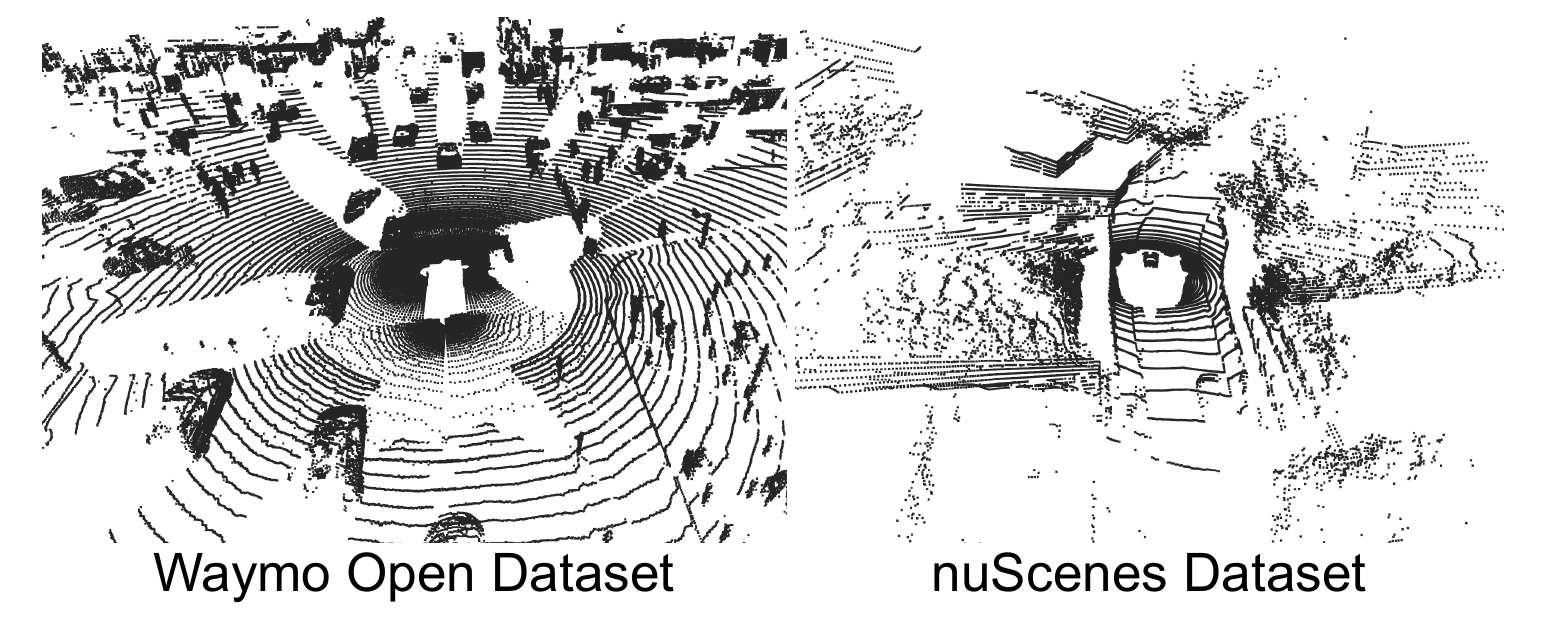}
\caption{Point Cloud Data Demonstration for Each Dataset.}
\label{fig:result_section_dataset_demo}
\end{figure}

To validate the performance of the proposed IGO-PQA algorithm and transformer-based regression model, we evaluate our work on both the nuScenes \cite{Caesar2019nuScenesAM} and Waymo \cite{Sun_2020_CVPR} open datasets, two of the most popular large-scale open-source datasets for autonomous driving perception studies.
For the nuScenes dataset, we use the official training and validation splits (28,130 training samples and 6,019 validation samples). The IGO-PQA is not evaluated on the nuScenes test samples due to the lack of public ground-truth labels. 
The input LiDAR point cloud data for nuScenes is collected from a single 32-beam LiDAR sensor.
For the Waymo open dataset, we create custom training and validation splits, each consisting of 99,034 samples. 
The input LiDAR point cloud data for Waymo is a fused point cloud collected from five LiDAR sensors. 
Figure \ref{fig:result_section_dataset_demo} presents a demonstration of the LiDAR point clouds from both the nuScenes and Waymo datasets.

The implementation of the IGO-PQA paradigm involves three steps:
(a) Construct the IGO-PQA score for both the training and validation samples based on the LiDAR point clouds and ground truth object labels.
(b) Train the IGO-PQA regression model using the training dataset, with point cloud data as input and the IGO-PQA score as the target.
(c) Evaluate the performance of the IGO-PQA regression model on the validation dataset.

For implementation details, the IGO-PQA regression model is trained on an NVIDIA A100-80G GPU using a cyclic learning rate schedule and optimized with the standard weighted Adam algorithm.
In addition, we open-source our work to further illustrate our implementation.

\subsection{IGO-PQA Generated Results Discussions}
For the proposed IGO-PQA generation algorithm, we evaluate our algorithm from two perspectives: (a) qualitative visualization of the IGO-PQA's generated results on both Waymo and nuScenes dataset, and (b) quantitative object detection algorithms' performance correlation with the IGO-PQA's generate quality score in the nuScenes dataset.
\subsubsection{IGO-PQA Score Visualization}
\begin{figure}[ht]
\centering
\includegraphics[width=8.5cm]{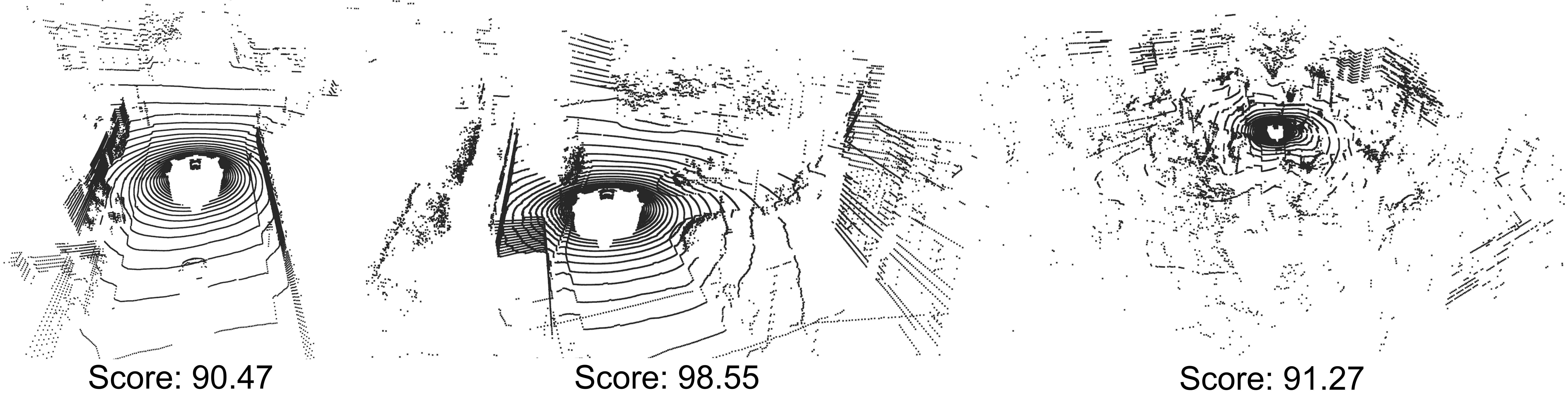}
\caption{nuScenes Dataset Visualization with High Quality.}
\label{fig:result_section_good_nuscenes}
\end{figure}

\begin{figure}[ht]
\centering
\includegraphics[width=8.5cm]{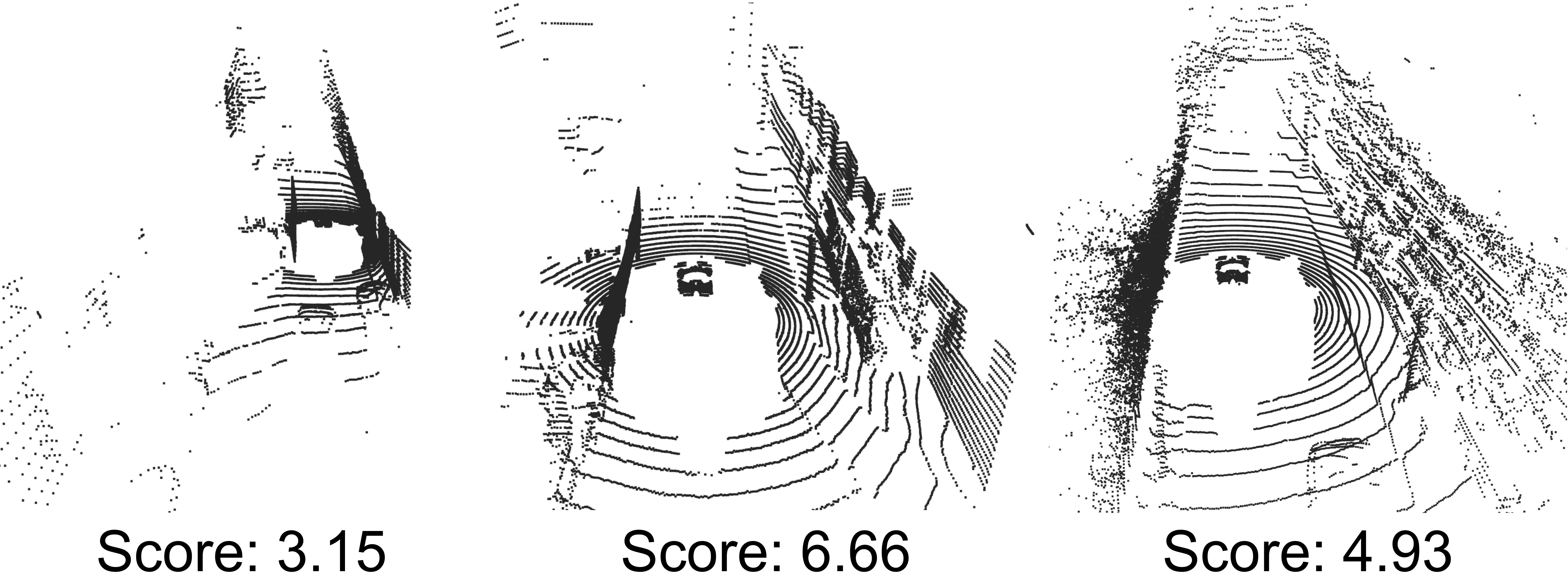}
\caption{nuScenes Dataset Visualization with Low Quality.}
\label{fig:result_section_bad_nuscenes}
\end{figure}

Figures \ref{fig:result_section_good_nuscenes} and \ref{fig:result_section_bad_nuscenes} showcase several examples of the IGO-PQA algorithm's results on the nuScenes dataset. 
These results illustrate that the distance captured in the point cloud is a crucial factor for achieving a high point cloud quality index.

For instance, in examples 2 and 3 of Figure \ref{fig:result_section_good_nuscenes}, the point clouds are not only relatively dense but also have a long detection range. 
Conversely, in the low-quality score examples (Figure \ref{fig:result_section_bad_nuscenes}), the point clouds are captured under conditions with a relatively narrow lateral distance and a far longitudinal distance, resulting in fewer captured points. Additionally, the content of the captured point cloud also affects quality. 
In example 3 of Figure \ref{fig:result_section_bad_nuscenes}, although the point density is high, most points represent objects such as walls, leading to a low point cloud perception quality.

\begin{figure}[ht]
\centering
\includegraphics[width=8.5cm]{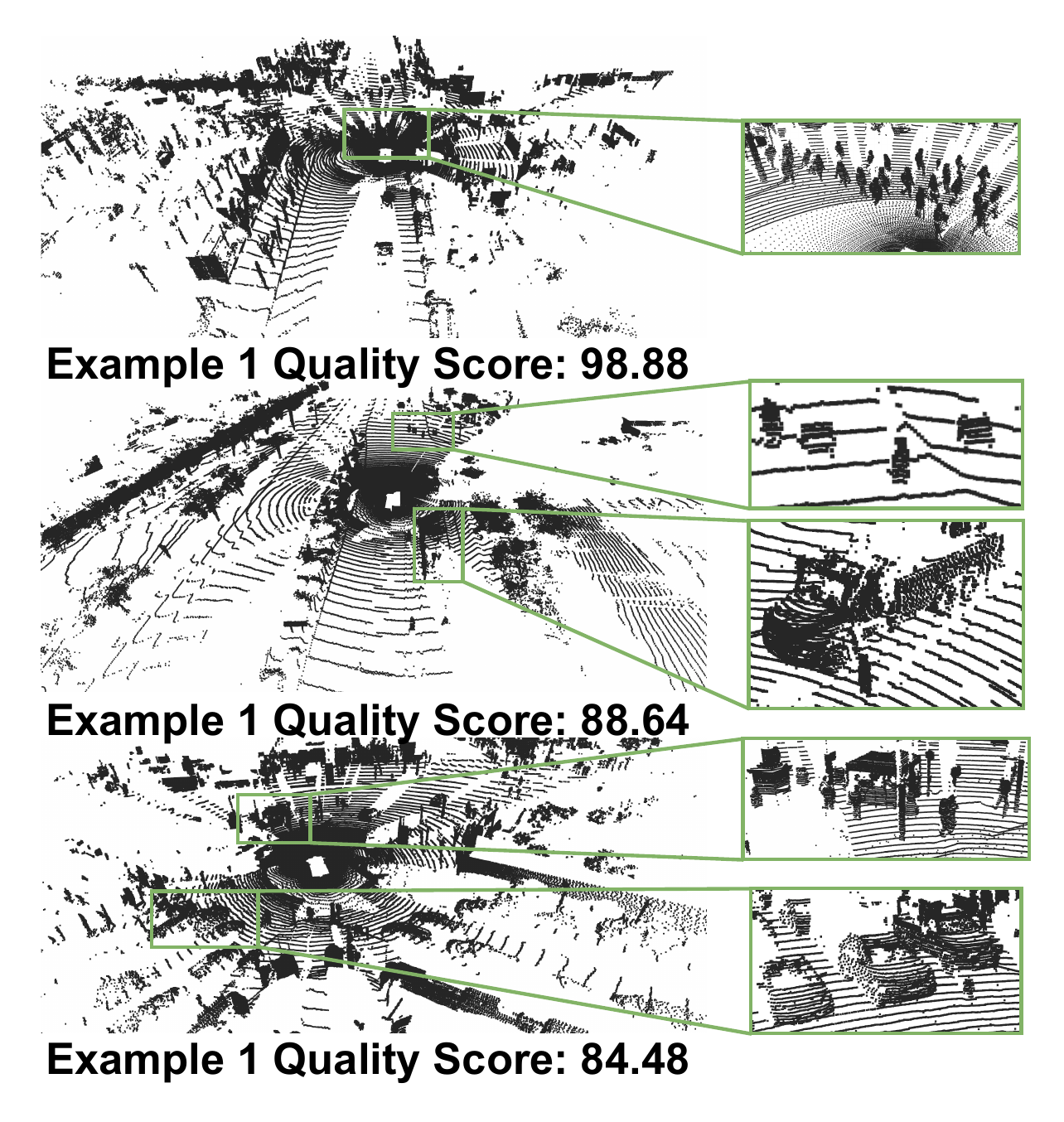}
\caption{Waymo Open Dataset Visualization with High Quality.}
\label{fig:result_section_good_waymo}
\end{figure}

\begin{figure}[ht]
\centering
\includegraphics[width=8.5cm]{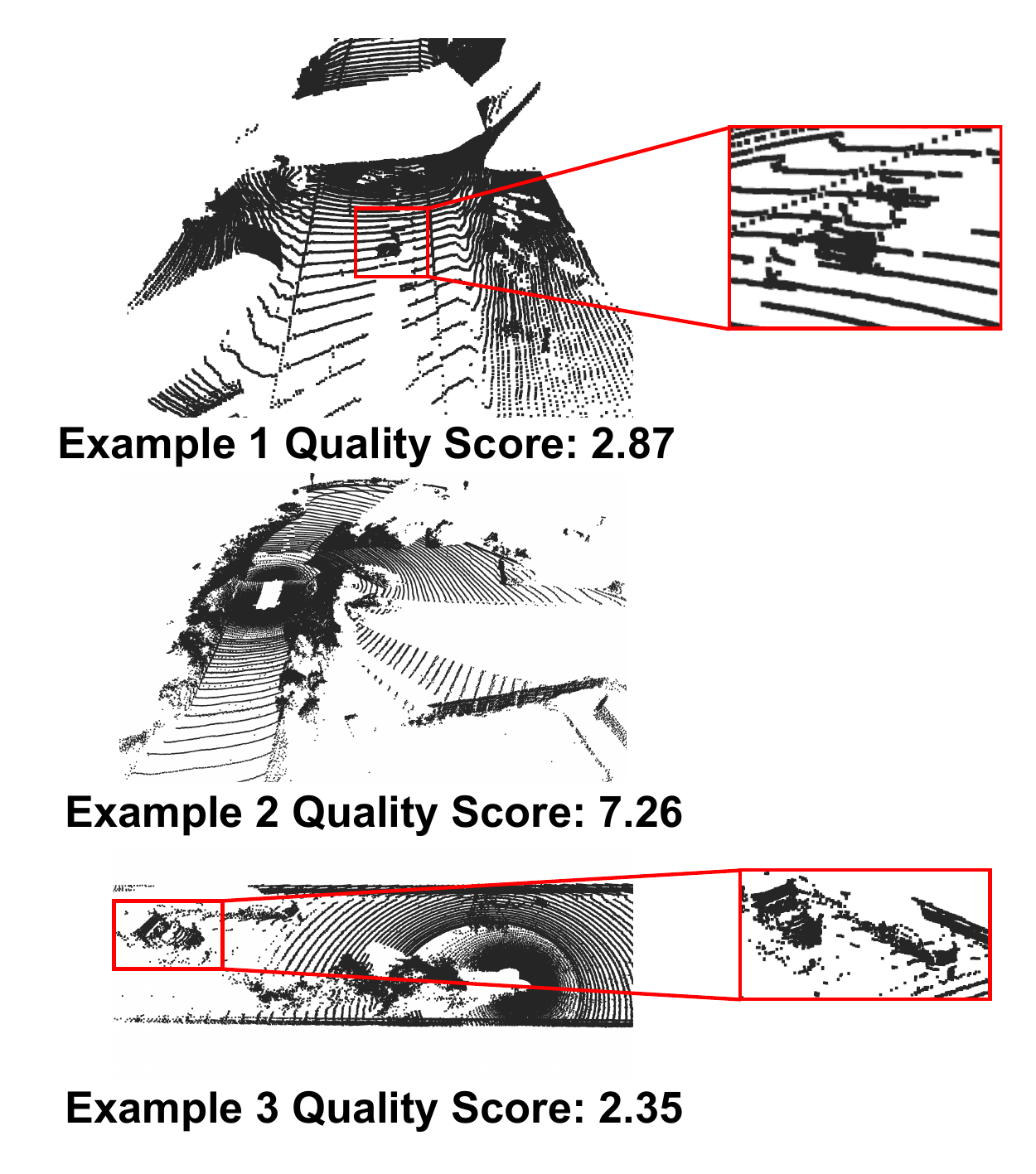}
\caption{Waymo Open Dataset Visualization with Low Quality.}
\label{fig:result_section_bad_waymo}
\end{figure}

Figures \ref{fig:result_section_good_waymo} and \ref{fig:result_section_bad_waymo} present several visualization examples from the Waymo Open dataset. 
High IGO-PQA scores in this dataset are associated with dense LiDAR point clouds and a long detection range, as seen in example 1 of Figure \ref{fig:result_section_good_waymo}. 
Moreover, objects like vehicles and pedestrians, as well as off-road agents, are captured clearly, as demonstrated in example 3 of Figure \ref{fig:result_section_good_waymo}. 
Similar to the nuScenes dataset, poor IGO-PQA scores in the Waymo dataset occur under conditions of narrow roads and poor capture of objects like traffic signs and road signs.

In summary, our qualitative observations of the proposed IGO-PQA algorithm's results indicate that the algorithm effectively captures the importance of good LiDAR-based point cloud quality, which includes factors such as detection range and high object density.

\subsubsection{Object Detection Algorithm Detection Performance Correlation}
Besides evaluating the proposed IGO-PQA from simple point cloud observation, we also utilize object detection algorithms detection results as a factor to examine our proposed IGO-PQA.
It is worth noting that this experiment is conducted in the nuScenes dataset, as most of the object detection algorithms are trained and evaluated at the nuScenes dataset and it is relatively easy to obtain the trained model weights and parameters.
For object detection algorithm evaluation, we separate the nuScenes dataset samples based on the IGO-PQA score, where 0$\sim$33.9 represents the "low-quality samples", "34.0$\sim$66.9" represents "medium-quality samples", and 67$\sim$100 represents the "high-quality samples".
After that, several object detection algorithms are employed to detect objects for the three categories, and their detection performances are used as proof of the correctness of the IGO-PQA score generation.
To ensure the detection performance is not biased, we utilize four 3D object detection algorithms: PointPillar, CenterPoint, TransFusion with LiDAR-only input, and VISTA.
The PointPillar and CenterPoint are CNN-based object detection algorithms that have been developed years ago, which are reliable and well-tested for autonomous driving applications.
The TransFusion and VISTA algorithms are transformer-based that are recently developed with better detection performance on different open-source datasets.

The results of the object detection performance among different quality samples are shown in Table \ref{table:result_section_object_detection_table}.
According to Table \ref{table:result_section_object_detection_table}, the average precision (AP) and F1-score are increasing with better quality samples for most objects.
This proves that the proposed IGO-PQA score can describe the overall point cloud quality with a given LiDAR frame.

\begin{table*}
\centering
\caption{Object Detection Results with Different IGO-PQA Categories}
\scalebox{0.85}{
\begin{tabular}{M|AAA|AAA|AAA|AAA}
\toprule
\multicolumn{13}{c}{\textbf{VISTA}}                                                                                                                                                                                                                                                                                                               \\\toprule
\textbf{Classes}          & \multicolumn{3}{c|}{\textbf{Car}}                                            & \multicolumn{3}{c|}{\textbf{Pedestrian}}                                     & \multicolumn{3}{c|}{\textbf{Motorcycle}}                                     & \multicolumn{3}{c}{\textbf{Barrier}}                                        \\\midrule
\textbf{IGO-PQA Category} & \textbf{Average Precision} & \textbf{F1-score} & \textbf{Num of Objects} & \textbf{Average Precision} & \textbf{F1-score} & \textbf{Num of Objects} & \textbf{Average Precision} & \textbf{F1-score} & \textbf{Num of Objects} & \textbf{Average Precision} & \textbf{F1-score} & \textbf{Num of Objects} \\\midrule
High                      & 0.680                      & 0.731             & 1,568                      & 0.820                      & 0.821             & 559                        & 0.586                      & 0.663             & 126                        & 0.667                      & 0.667             & 502                        \\\midrule
Medium                    & 0.648                      & 0.703             & 49,079                     & 0.705                      & 0.729             & 24,599                     & 0.603                      & 0.644             & 3,857                      & 0.555                      & 0.632             & 18,734                     \\\midrule
Low                       & 0.612                      & 0.680             & 29,357                     & 0.671                      & 0.703             & 9,189                      & 0.449                      & 0.594             & 1,132                      & 0.524                      & 0.600             & 7,756                      \\\toprule
\multicolumn{13}{c}{\textbf{TransFusion-LiDAR}}                                                                                                                                                                                                                                                                                                   \\\toprule
\textbf{Classes}          & \multicolumn{3}{c|}{\textbf{Car}}                                            & \multicolumn{3}{c|}{\textbf{Pedestrian}}                                     & \multicolumn{3}{c|}{\textbf{Motorcycle}}                                     & \multicolumn{3}{c}{\textbf{Barrier}}                                        \\\midrule
\textbf{IGO-PQA Category} & \textbf{Average Precision} & \textbf{F1-score} & \textbf{Num of Objects} & \textbf{Average Precision} & \textbf{F1-score} & \textbf{Num of Objects} & \textbf{Average Precision} & \textbf{F1-score} & \textbf{Num of Objects} & \textbf{Average Precision} & \textbf{F1-score} & \textbf{Num of Objects} \\\midrule
High                      & 0.716                      & 0.762             & 1,568                      & 0.684                      & 0.725             & 559                        & 0.704                      & 0.793             & 126                        & 0.486                      & 0.619             & 502                        \\\midrule
Medium                    & 0.657                      & 0.727             & 49,079                     & 0.576                      & 0.644             & 24,599                     & 0.650                      & 0.725             & 3,857                      & 0.468                      & 0.594             & 18,734                     \\\midrule
Low                       & 0.621                      & 0.697             & 29,357                     & 0.540                      & 0.615             & 9,189                      & 0.635                      & 0.713             & 1,132                      & 0.459                      & 0.519             & 7,756                      \\\toprule
\multicolumn{13}{c}{\textbf{PointPillar}}                                                                                                                                                                                                                                                                                                         \\\toprule
\textbf{Classes}          & \multicolumn{3}{c|}{\textbf{Car}}                                            & \multicolumn{3}{c|}{\textbf{Pedestrian}}                                     & \multicolumn{3}{c|}{\textbf{Motorcycle}}                                     & \multicolumn{3}{c}{\textbf{Barrier}}                                        \\\midrule
\textbf{IGO-PQA Category} & \textbf{Average Precision} & \textbf{F1-score} & \textbf{Num of Objects} & \textbf{Average Precision} & \textbf{F1-score} & \textbf{Num of Objects} & \textbf{Average Precision} & \textbf{F1-score} & \textbf{Num of Objects} & \textbf{Average Precision} & \textbf{F1-score} & \textbf{Num of Objects} \\\midrule
High                      & 0.594                      & 0.660             & 1,568                      & 0.538                      & 0.596             & 559                        & 0.122                      & 0.257             & 126                        & 0.057                      & 0.192             & 502                        \\\midrule
Medium                    & 0.566                      & 0.637             & 49,079                     & 0.407                      & 0.493             & 24,599                     & 0.067                      & 0.174             & 3,857                      & 0.081                      & 0.196             & 18,734                     \\\midrule
Low                       & 0.549                      & 0.634             & 29,357                     & 0.342                      & 0.431             & 9,189                      & 0.085                      & 0.204             & 1,132                      & 0.068                      & 0.182             & 7,756                      \\\toprule
\multicolumn{13}{c}{\textbf{CenterPoint}}                                                                                                                                                                                                                                                                                                         \\\toprule
\textbf{Classes}          & \multicolumn{3}{c|}{\textbf{Car}}                                            & \multicolumn{3}{c|}{\textbf{Pedestrian}}                                     & \multicolumn{3}{c|}{\textbf{Motorcycle}}                                     & \multicolumn{3}{c}{\textbf{Barrier}}                                        \\\midrule
\textbf{IGO-PQA Category} & \textbf{Average Precision} & \textbf{F1-score} & \textbf{Num of Objects} & \textbf{Average Precision} & \textbf{F1-score} & \textbf{Num of Objects} & \textbf{Average Precision} & \textbf{F1-score} & \textbf{Num of Objects} & \textbf{Average Precision} & \textbf{F1-score} & \textbf{Num of Objects} \\\midrule
High                      & 0.666                      & 0.723             & 1,568                      & 0.717                      & 0.762             & 559                        & 0.557                      & 0.612             & 126                        & 0.471                      & 0.487             & 502                        \\\midrule
Medium                    & 0.624                      & 0.696             & 49,079                     & 0.597                      & 0.664             & 24,599                     & 0.473                      & 0.543             & 3,857                      & 0.442                      & 0.576             & 18,734                     \\\midrule
Low                       & 0.592                      & 0.675             & 29,357                     & 0.567                      & 0.637             & 9,189                      & 0.462                      & 0.525             & 1,132                      & 0.424                      & 0.532             & 7,756      \\\bottomrule               
\end{tabular}}
\label{table:result_section_object_detection_table}
\end{table*}
\subsection{IGO-PQI Regression Model Results Discussions}
This section discusses the proposed transformer-based point cloud regression neural network model. It is worth noticing that since our work is by far a novel task for outdoor LiDAR point cloud quality assessment, there are hardly any baseline models or benchmarks that we can compare. Therefore, we compare our proposed transformer-based model with the two baseline models that we created.

Table \ref{table:results_regression_nuscenes} presents the performance of the IGO-PQA regression model, and the voxel-based transformer model is observed to exhibit the best performance, with a 2.5\% improvement over the baseline model. This finding highlights the importance of incorporating a transformer-based model in the regression process to effectively capture the spatial relationships and dependencies of the point cloud data.

\begin{figure}[ht]
\centering
\includegraphics[width=8.5cm]{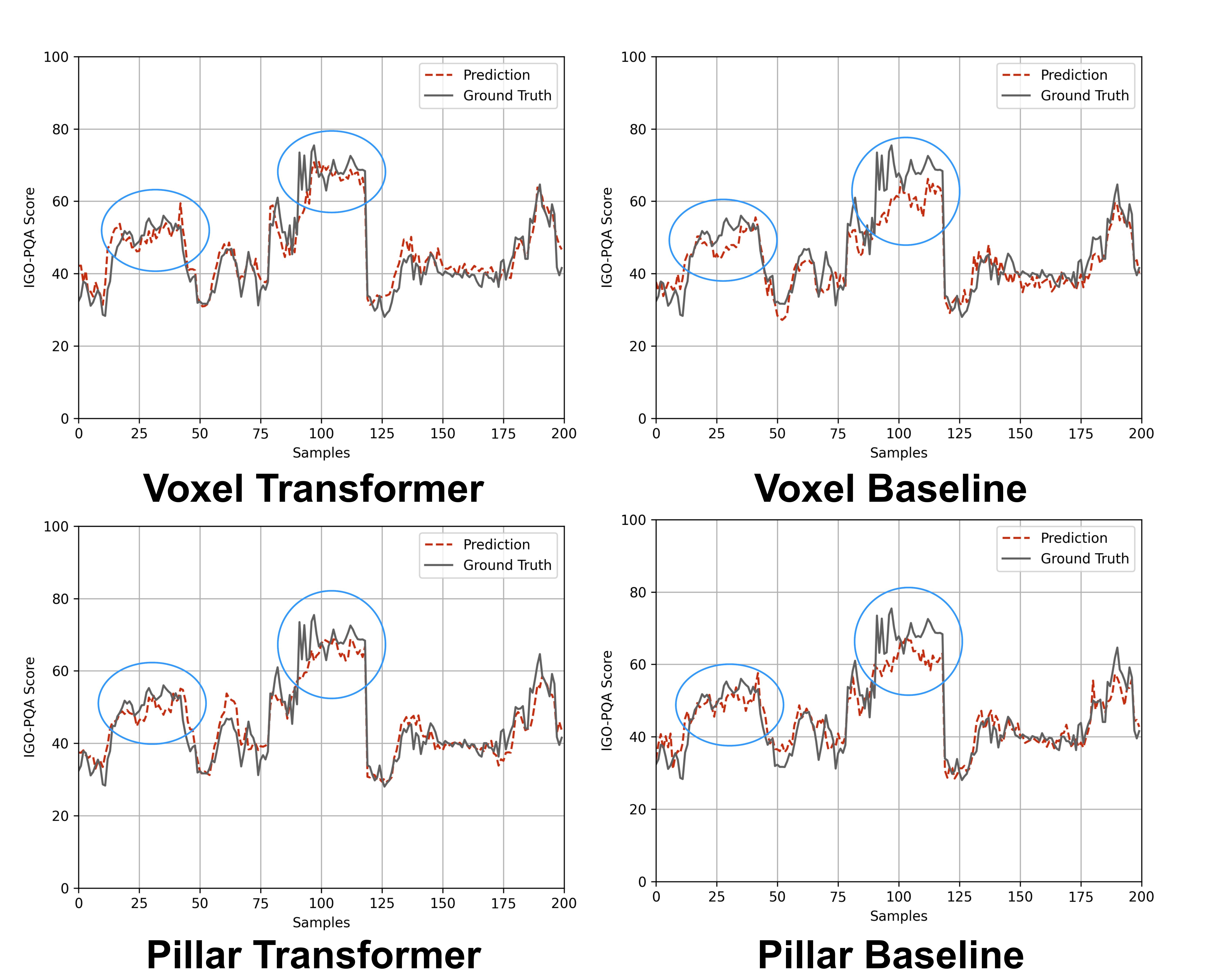}
\caption{nuScenes Dataset LiDAR Quality Index Regression Demonstration.}
\label{fig:result_section_lqa_demo_nuscenes}
\end{figure}

\begin{figure}[ht]
\centering
\includegraphics[width=8.5cm]{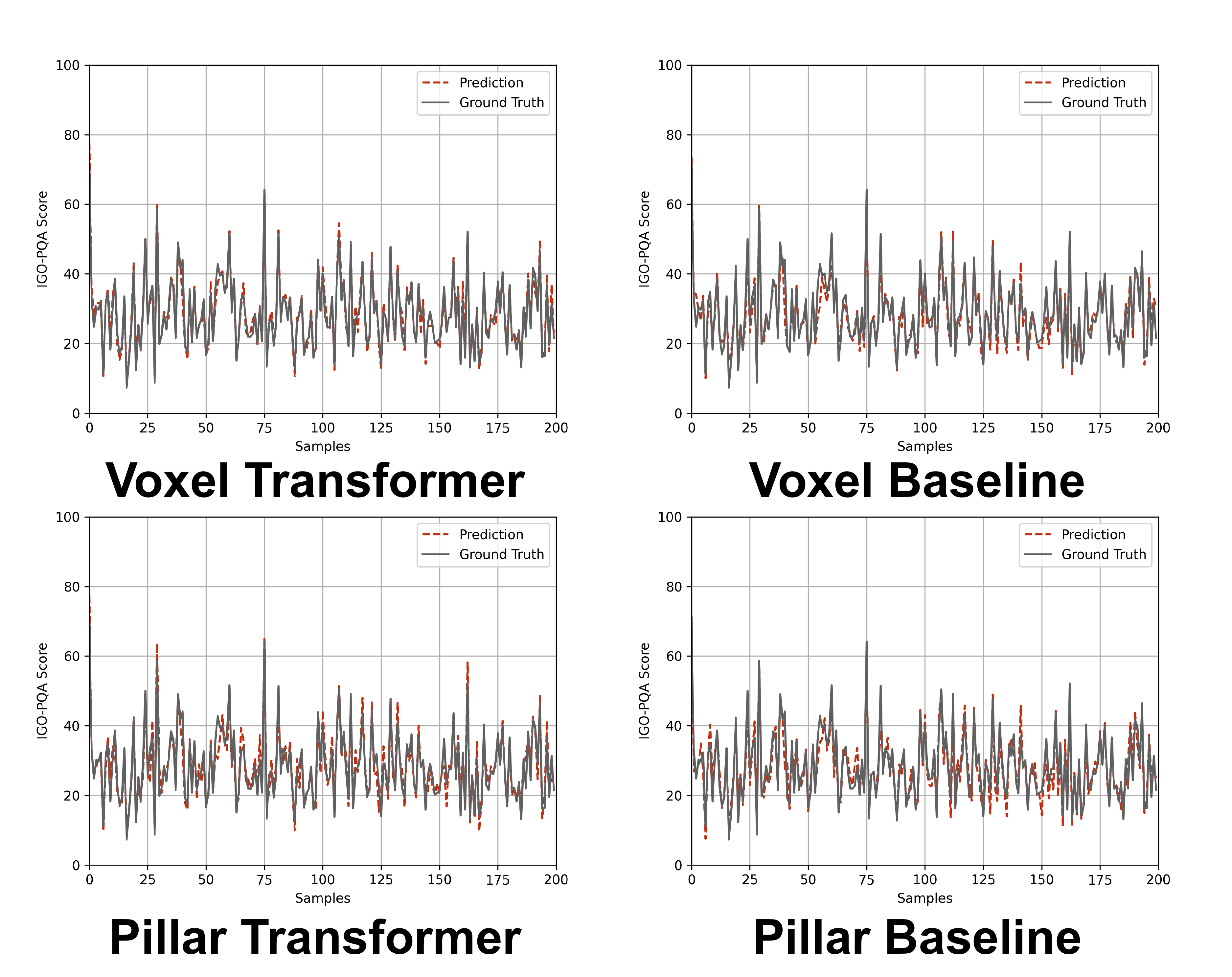}
\caption{Waymo Open Dataset LiDAR Quality Index Regression Demonstration.}
\label{fig:result_section_lqa_demo_waymo}
\end{figure}

\begin{table}
\centering
\caption{IGO-PQA Regression Results on nuScenes Dataset}
\begin{tabular}{L|M|M|M|M}
\toprule
                            & {\color[HTML]{CB0000} \textbf{Voxel Former}}      & \textbf{Voxel Baseline} & \textbf{Pillar Former} & \textbf{Pillar Baseline} \\\midrule
\textbf{PLCC}               & {\color[HTML]{CB0000} \textbf{0.864}}            & 0.855                   & 0.846                  & 0.843                     \\\midrule
\textbf{SRCC}               & {\color[HTML]{CB0000} \textbf{0.876}}            & 0.865                   & 0.854                  & 0.857                     \\\midrule
\textbf{Avg. L1}      & {\color[HTML]{CB0000} \textbf{4.32}}             & 4.51                    & 4.52                   & 4.55                      \\\bottomrule
\end{tabular}
\label{table:results_regression_nuscenes}
\end{table}

Figure \ref{fig:result_section_lqa_demo_nuscenes} presents a comparison between the model-estimated IGO-PQA and the ground truth, with the results indicating that the proposed IGO-PQA regression model's results are closer to the ground truth. These results demonstrate the potential of the IGO-PQA approach for accurately assessing the quality of LiDAR-based point clouds, particularly in terms of facilitating object detection algorithms for autonomous driving.

For the Waymo Open Dataset, the results exhibit a similar pattern to the nuScenes dataset, with the voxel-based transformer regression network demonstrating the best performance, as shown in \ref{table:results_regression_waymo} and \ref{fig:result_section_lqa_demo_waymo}. 
Notably, the regression performance on the Waymo Open Dataset is significantly better than that on the nuScenes dataset. 
We attribute this improvement to Waymo's training sample set being nearly four times larger than that of the nuScenes dataset, leading to enhanced performance. 

\begin{table}
\centering
\caption{IGO-PQA Regression Results on Waymo Open Dataset}
\begin{tabular}{L|M|M|M|M}
\toprule
                            & {\color[HTML]{CB0000} \textbf{Voxel Former}}     & \textbf{Voxel Baseline} & \textbf{Pillar Former} & \textbf{Pillar Baseline}  \\\midrule
\textbf{PLCC}               & {\color[HTML]{CB0000} \textbf{0.975}}            & 0.950                   & 0.921                  & 0.910                       \\\midrule
\textbf{SRCC}               & {\color[HTML]{CB0000} \textbf{0.970}}            & 0.941                   & 0.903                  & 0.890                       \\\midrule
\textbf{Avg. L1}            & {\color[HTML]{CB0000} \textbf{1.641}}            & 2.406                   & 3.008                  & 3.183                       \\\midrule
\textbf{Runtime (s)}    & {\color[HTML]{CB0000} \textbf{9.1}}              & 9.6                     & 13.5                   & 14.8                       \\\bottomrule
\end{tabular}
\label{table:results_regression_waymo}
\end{table}

\begin{table}
\centering
\caption{Ablation Study with the Transformer's Number of Decoder Layers}
\begin{tabular}{L|L|L|L}
\toprule
\textbf{Decoder Layer}   & \textbf{PLCC}    & \textbf{SRCC}     & \textbf{Mean L1 Error}     \\\toprule
\textbf{1}               & 0.856            & 0.869             & 4.41                       \\\midrule
\textbf{2}               & 0.864            & 0.876             & 4.32                       \\\midrule
\textbf{3}               & 0.868            & 0.874             & 4.34                       \\\midrule
\textbf{4}               & 0.862            & 0.874             & 4.36                       \\\midrule
\textbf{5}               & 0.861            & 0.871             & 4.35                       \\\midrule
\textbf{6}               & 0.859            & 0.869             & 4.39                       \\\bottomrule
\end{tabular}
\label{table:results_ablation_1}
\end{table}

\begin{table}
\centering
\caption{Ablation Study with the Different Patch Sizes}
\begin{tabular}{M|M|M|M}
\toprule
\textbf{Patch Size} & \textbf{PLCC} & \textbf{SRCC} & \textbf{Mean L1 Error} \\\toprule
\textbf{4}          & 0.861         & 0.873         & 4.38                   \\\midrule
\textbf{8}          & 0.864         & 0.876         & 4.32                   \\\midrule
\textbf{16}         & 0.858         & 0.863         & 4.40                   \\\midrule
\textbf{32}         & 0.854         & 0.861         & 4.43                   \\\bottomrule
\end{tabular}
\label{table:results_ablation_2}
\end{table}

\begin{table}[ht]
\centering
\caption{Ablation Study with the Different Positional Encoding Methods}
\begin{tabular}{M|M|M|M}
\toprule
\textbf{Positional Encoding} & \textbf{PLCC} & \textbf{SRCC} & \textbf{Mean L1 Error} \\\toprule
N/A                          & 0.759         & 0.771         & 4.96                   \\\midrule
Sinusoidal                   & 0.864         & 0.876         & 4.32                   \\\midrule
Learned                      & 0.866         & 0.874         & 4.33                   \\\bottomrule
\end{tabular}
\label{table:results_ablation_3}
\end{table}

\begin{figure}[!ht]
\centering
\includegraphics[width=8.5cm]{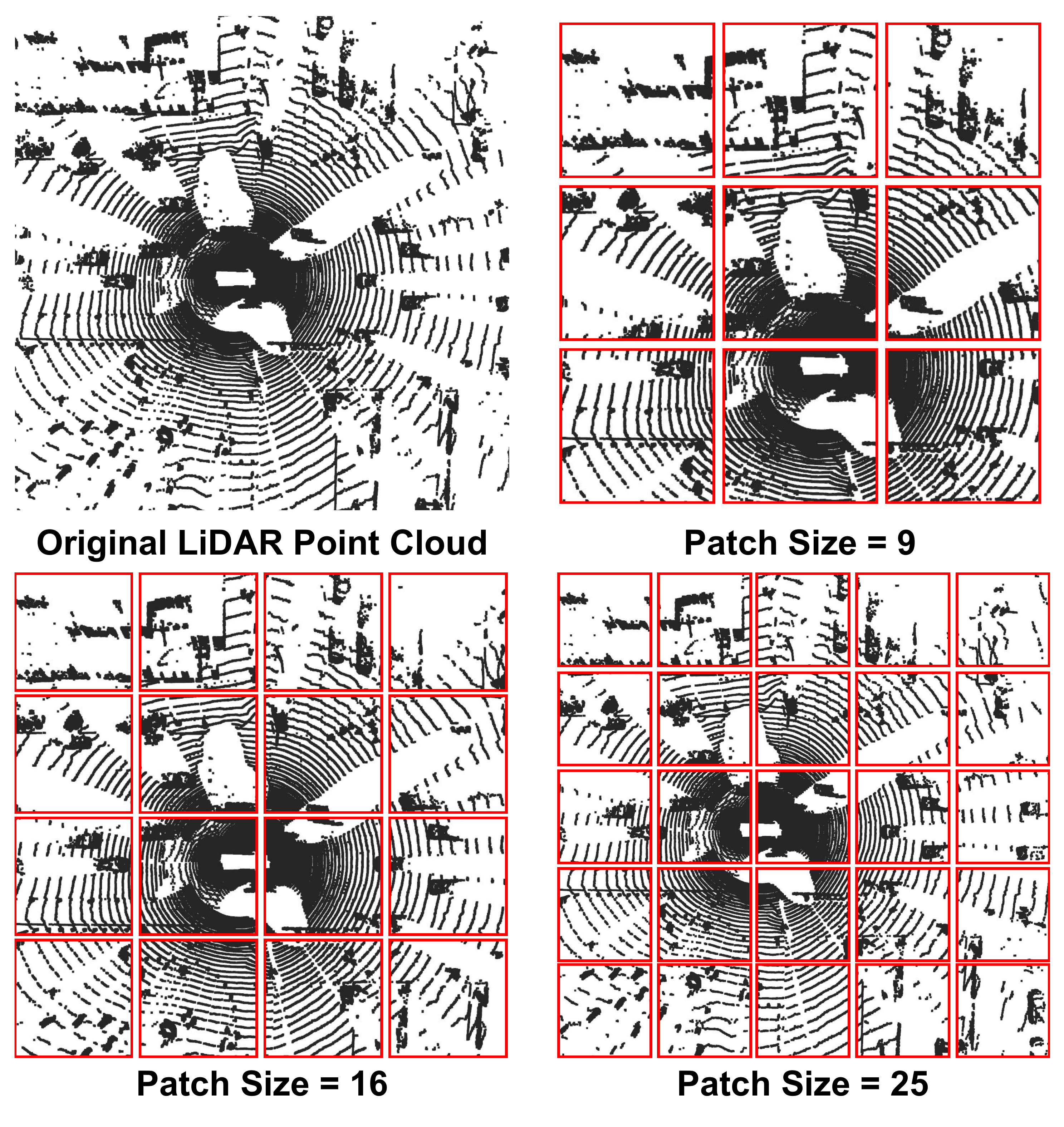}
\caption{Demonstration of Ablation Studies on Different Patch Size}
\label{fig:results_patch}
\end{figure}

\subsection{Ablation Studies}
We have conducted extensive ablation studies to prove the effectiveness and robustness of our proposed transformer-based regression module, which includes the number of transformer layers, feature map patch size, and positional encoding methods.

Table \ref{table:results_ablation_1} indicates that the regression performance of the transformer decoder layer is optimized when the number of layers is set to 2. 
However, this finding differs from the results of the DETR-based detection ablation study, where an increase in the number of decoder layers can enhance detection performance.
The reason behind this discrepancy is that the regression task is simpler than the detection task since the model only needs to learn a single index, unlike the detection model which must learn various aspects such as object localization, classes, and object sizes. 
As a result, increasing the number of decoder layers may cause the model to overfit, which can negatively impact regression performance.

Table \ref{table:results_ablation_2} presents the regression performance with increasing patch sizes. 
The results indicate that the regression performance is highest when the patch size is 8. 
To gain a better understanding of patch size, Figure \ref{fig:results_patch} illustrates the concept of increasing patch sizes. 
Since the LiDAR feature map is sparser than an image feature map, increasing the patch size results in empty feature patches. 
Consequently, while increasing the patch size may improve the feature extraction resolution, it does not necessarily enhance performance.

Table \ref{table:results_ablation_3} presents the ablation results obtained with different positional encoding methods. According to the findings, the positional encoding of the feature map patches and point cloud is crucial to the IGO-PQA regression model. 
However, there is no significant difference between the conventional sinusoidal and learned-based positional encoding methods. This outcome is consistent with the conclusions drawn from ViT studies.

\section{Conclusions}
This paper delves into the evaluation of LiDAR-based point clouds for their use in autonomous driving. 
To this end, a new approach called IGO-PQA is introduced, which leverages an image-based saliency map as a reference to generate point-based saliency intensity. 
This method offers a precise assessment of point cloud quality, especially in terms of supporting object detection algorithms. 
Furthermore, the proposed IGO-PQA regression model can generate highly accurate IGO-PQA scores without relying on ground truth labels, thus making it well-suited for online, real-time applications. 
The proposed regression model yields a PLCC of 0.86 and an SRCC of 0.87 for regression scores in the nuScenes dataset and PLCC of 0.98 and an SRCC of 0.97 for the Waymo open dataset.

Despite the consistent quality assessment scores and accurate regression performance, we recognize the limitations and future research topics based on our current findings. 
Firstly, the proposed transformer-based regression model lacks comparison with benchmark models. 
To the best of our knowledge, there is no non-reference LiDAR-based point cloud quality regression model available for comparison. 
As a result, we conducted extensive ablation studies that evaluated two datasets to establish a benchmark for future researchers.
Secondly, the application of the proposed perception quality score to online perception tasks to enhance performance remains an open question. 
For instance, future research could explore how to integrate our point cloud perception quality score with object detection or segmentation algorithms to improve perception performance. 
Another possibility is using the perception quality score as a guideline for planning algorithms, leading to more robust and safe trajectory generation under varying perception quality conditions.

We believe that analyzing and quantifying point cloud quality is a crucial step toward gaining a deeper understanding of LiDAR-based point clouds for 3D perception in autonomous driving. 
By systematically evaluating the quality of point cloud data, we can ultimately enhance the reliability and accuracy of autonomous vehicle perception systems, which contributes to the development of safer and more efficient autonomous driving technologies.
\bibliography{literature_reviews}

\end{document}